# Can LLMs assist with Ambiguity? A Quantitative Evaluation of various Large Language Models on Word Sense Disambiguation

T.G.D.K.Sumanathilaka, Nicholas Micallef, Julian Hough

Department of Computer Science, Swansea University, Wales, UK
{t.g.d.sumanathilaka, nicholas.micallef, julian.hough }@swansea.ac.uk

**Abstract**

Ambiguous words are often found in modern digital communications. Lexical ambiguity challenges traditional Word Sense Disambiguation (WSD) methods, due to limited data. Consequently, the efficiency of translation, information retrieval, and question-answering systems is hindered by these limitations. This study investigates the use of Large Language Models (LLMs) to improve WSD using a novel approach combining a systematic prompt augmentation mechanism with a knowledge base (KB) consisting of different sense interpretations. The proposed method incorporates a human-in-loop approach for prompt augmentation where prompt is supported by Part-of-Speech (POS) tagging, synonyms of ambiguous words, aspect-based sense filtering and few-shot prompting to guide the LLM. By utilizing a few-shot Chain of Thought (COT) prompting-based approach, this work demonstrates a substantial improvement in performance. The evaluation was conducted using FEWS test data and sense tags. This research advances accurate word interpretation in social media and digital communication.

## 1 Introduction

In Natural Language Processing (NLP), identifying the exact meaning of words within sentences is key. This is because misunderstandings of word sense can lead to false information which results in misinformation. In the context of Cyber Threat Intelligence, such misinformation and ambiguity can conceal the true nature of threats, leading to inadequate responses and potentially leaving systems vulnerable (Arazzi et al., 2023). Words that have multiple meanings (polysemy) are a major challenge that NLP can overcome using computational methods. Even though there's been a lot of research on figuring out the right meaning of words (WSD) in different languages, using various methods, it has not been completely successful (Mente et al., 2022). For instance, previous studies performed on WSD have not been able to solve some tricky cases due to its poor contextual understanding by the models (Nguyen et al., 2018). However, research shows that a word's meaning is closely linked to the words around it proving that isolated word analysis is insufficient to perform correct sense identification (Luo et al., 2018). Therefore, proper word sense with positional value, POS tag and aspect of the sentence is being considered for accurate models. LLMs and generative AI, which are based on transformers, show promising results in the contextual understanding of words (Dettmers et al., 2023). These models have shown a strong ability to handle complex language tasks because of extensive training on vast amounts of data. Finetuning such base models for downstream tasks such as question answering and domain specific knowledge generation has shown promising results (Guo et al., 2023). In our study, we were mainly focused on evaluating how LLMs can be used for specific downstream tasks like WSD by investigating their capability of identifying the right meaning of words. More specifically, we want to understand if LLMs can be used to match words with multiple meanings to their correct sense in a sentence. Even if research focuses on supervised WSD methods (often using paradigmatic relationships like synonyms, hyponyms, and hypernyms), this study explores an alternative path to ensemble different computational techniques like KB to improve the prediction accuracy for WSD. Prior work on WSD has attempted to extract the correct gloss/sense tag by reframing the problem into different aspects. However, the major limitation of these studies is identifying a sense of tricky instances with diverse sense meaning distribution. Instances of lexical ambiguity, such as words like 'post', 'brake', 'part', 'bat' and 'try' exhibit significant diversity, with each possessing more than ten distinct senses across both noun and verb forms. Pasini *et al.* have found that current proposed architectures are not confident enough to predict the sense for highly diverse ambiguous words, where there are multiple interpretations for the ambiguous word (Pasini et al., 2021). Drawing upon insights from existing literature, our work aims to evaluate the impact of using pre-trained language models for sense prediction for diverse ambiguous words and to identify the key factors influencing the performance of sense prediction of highly ambiguous words. To overcome the above limitation, we use the pretrained knowledge of the language models as these models (PLMs) are trained on massive amounts of text, offering a potential for addressing this data scarcity issue in supervised learning. We have evaluated multiple pipelines to measure LLM capabilities with commercial LLM like GPT 3.5 Turbo, GPT 4 Turbo and Gemini and Open Source models like Gemma 7B (Gemma Team et al., 2024), Mixtral (Jiang et al., 2024), Llama-2-70B (Touvron et al., 2023), Llama-3-70B, Yi 34B (AI et al., 2024). WSD pipelines have been evaluated with simple prompts, sentence augmentation for improved understanding, and a hybrid Retrieval Augmented Generation (RAG) inspired model that blends the LLM with a KB. This process follows a human-in-loop approach and the identification of the optimal prompting technique to test the rest of the LLMs. The identified advanced prompting technique has been used to evaluate LLM's capabilities for WSD and the outcomes are presented in the results section.

The contributions of the study can be highlighted as follows.

- Performing a detailed evaluation of open source and commercial LLMs capabilities when handling lexical ambiguity.
- Incorporating and evaluating aspect-based sense filtering and use of synonyms to improve model performance when dealing with highly ambiguous words.

In summary, this introduction has given a thorough look at the field of research and the main points we'll be exploring. Going forward, the following sections will dive into similar studies, explain our chosen research methods, present our findings, and discuss the limitations of the proposed methods. We will also suggest areas for potential future research. The goal is to provide a proper understanding of this topic and make a meaningful contribution to the ongoing conversation.

## 2 Related works

Ambiguity in Natural language poses a significant challenge for various Natural Language Processing tasks, with WSD being a fundamental problem. WSD has been one of the continuing research areas in different languages as the proper word sense directly impacts many NLP tasks like machine translation, question answering, text summarization, text classification, and word sense induction. Several advanced new neural architectures have been suggested by many researchers for the WSD task by integrating KB models (Abeysiriwardana & Sumanathilaka, 2024). Different NLP techniques are grouped to perform the effective WSD task and an overview of these works can be found below.

### 2.1 Supervised WSD

Supervised approaches to WSD utilize labelled datasets to train models for sense disambiguation like Semcor, FEWS and Wordnet (Scarlini et al., 2020). Various algorithms and enhancements to the existing models have been proposed to enhance the accuracy of supervised WSD systems. For instance, the use of stacked bidirectional Long Short-Term Memory (LSTM) neural networks coupled with attention mechanisms has been explored (Laatar et al., 2023). This approach employed deep embedding-based representations of sentences containing ambiguous words, followed by self-attention mechanisms to highlight contextual features and construct overall semantic representations of sentences. Data augmentation techniques like Sense-Maintained Sentence Mixup (SMSMix) have also been introduced to increase the frequency of least frequent senses (LFS) and reduce distributional bias during training (Yoon et al., 2022). BiLSTM, which has shown promising results in detecting lexical ambiguities, particularly in low-resource languages (Le et al., 2018) and Enhanced WSD Integrating Synset Embeddings and Relations (EWISER) which integrates information from the LKB graph and pre-trained synset embeddings (Bevilacqua & Navigli, 2020) have been explored. The study GlossBERT improved the utilization of the gloss knowledge by constructing context gloss pairs reframing the WSD problem to sentence pair classification and presenting with three BERT based models (Huang et al., 2020). The nearby sense has been well used in some studies to outperform the predictions (Barba et al., 2021). Not only the above studies but also context dependent method (Koppula et al., 2021), multiple sense identification (Orlando et al., 2021), incorporating synonyms and example phrases (Song et al., 2021) have been used for the WSD task.

## 2.2 Knowledge base WSD

KB approaches to WSD utilize external resources like lexical databases and ontologies to clarify word senses. These methods employ semantic similarity measures and graph-based algorithms. For instance, a graph-based algorithm for Hindi WSD used Hindi WordNet to create weighted graphs representing word senses and their relations (Jha et al., 2023). Bootstrapping techniques integrating WordNet synsets have shown comparative improvement in WSD performance. Various KB approaches proposed innovative techniques for ambiguity resolution using semantic information. An adaptive sentence semantic similarity-based complex network approach represented ambiguous sentences as vertices, constructing a weighted complex network based on semantic similarities to resolve ambiguity. Context-aware semantic similarity measurement has enhanced unsupervised WSD by incorporating contextual information into similarity measurement, potentially improving model performance. Wang et al. introduce the Synset Relation-Enhanced Framework (SREF), expanding the WSD toolkit by augmenting basic sense embeddings with sense relations (Wang & Wang, 2020). Rouhizadeh et al. proposed a novel KB technique for Persian WSD, utilizing a pre-trained LDA model to assign ambiguous content words to topics and selecting the most probable sense based on similarity with FarsNet glosses. Additionally, studies have investigated semi-supervised WSD using graph-based SSL algorithms and various word embeddings combined with POS tags and word context. Cross-lingual approaches have also been explored, with investigations into cross-lingual word sense embedding and contextual word-level translation (Rudnick, 2011). Additionally, efforts in entity disambiguation proposed innovative formulations, such as ExtEnD, which frame the task as a text extraction problem and utilized transformer-based architectures to improve disambiguation accuracy (Barba et al., 2022). These approaches highlight the importance of considering linguistic diversity and resource availability in WSD research. Approaches like Sin-Sense pioneer cross-lingual sense disambiguation have used another language to aid the process in Sinhala WSD (Subasinghe, 2020). Furthermore, Sumanathilaka et al. proposed a suggestion level module to incorporate trie structure for Romanized Sinhala word prediction showing the importance of KB models (T. G. D. K. Sumanathilaka et al., 2023).

## 2.3 Hybrid approach with WSD

Hybrid methodologies emerge as promising avenues for WSD. TWE-WSD has incorporated a topical word embedding-based method integrating Latent Dirichlet Allocation (LDA) and word embedding techniques (Jia et al., 2021). However, it is important to note that approaches like TWE-WSD may have limitations when handling complex linguistic phenomena like homonymy (words with the same spelling but different meanings). Further, a study investigating English word translation versions using a hybrid strategy based on cyber translation aid and Wordnet 3.0 revealed different information demands for WSD, highlighting the importance of considering these nuances (Ji & Xiao, 2013).

## 2.4 Large language models for WSD

Sainz et al. demonstrate that LLMs have an inherent understanding of word senses, as evidenced by their ability to perform WSD without explicit training (Sainz et al., 2023). The authors achieved this by leveraging domain knowledge and associating words with specific fields like finance or biology. They frame WSD as a textual entailment problem, asking LLMs to determine if a domain label accurately describes a sentence containing an ambiguous word. Surprisingly, this zero-shot approach surpasses random guesses and sometimes rivals supervised WSD systems. This finding has been further supported by other empirical studies (Ortega-Martín et al., 2023). Additionally, cross-lingual word sense evaluation with contextual word-level translation on pre-trained language models has been investigated, and zero-shot WSD has been assessed using cross-lingual knowledge (Kang et al., 2023). Beyond prediction tasks, GPT-2 has been employed for contextual data augmentation, demonstrating the broad utility of LLMs in this field (Saidi et al., 2023). Research extends to areas like CLIP-based WSD for image retrieval (Pan et al., 2023) and language model analysis and evaluation (Loureiro et al., 2021), further exploring the capabilities of LLMs. This enhancement motivates the use of LLMs in our research to improve WSD.

## 3 Methodology

According to previous studies, it is evident that the usage of LLMs based approaches for WSD tasks can be effective. In our work, we evaluate the understanding of lexical ambiguity by different LLMs using different computational approaches like parameter tuning and prompt augmentation. The optimal prompt has been constructed using an iterative approach. The results were analyzed on corner cases and the augmented prompts have been tested and evaluated. The prompt augmentation process utilized different techniques like few-shot COT prompting and its variations.

### 3.1 Dataset selection

This work uses the FEWS dataset, which contains the sense tag list, training data and test data (Blevins et al., 2021). The selection of FEWS for the study mainly influenced it nature of the data, where it contains less frequently used ambiguous words compared to the Unified Evaluation framework (Raganato et al., 2017). In all the proposed approaches, the models were evaluated for their ability to correctly assign sense tags to ambiguous words positioned between <WSD> tokens within sentences. The sense tag definition from the FEWS sense tag is shown in Table 1.

| *Sense_id: | dictionary. noun.0 | Tags | en |
|---|---|---|---|
| *Word | dictionary | Depth | 1 |
| *Gloss | A reference work with a list of words from one or more languages, normally ordered alphabetically, explaining each word's meaning, and sometimes containing information on its etymology, pronunciation, usage, translations, and other data. | | |
| *Synonyms | wordbook | | |

\* Model Parameters used for the study.

Table 1: Sense tag definition.

The input sentence and the expected output are shown in Table 2.

| Input sentence | The aspiring author meticulously cross-checked her manuscript against various <WSD> dictionaries </WSD>, striving to ensure both word choice and proper usage. |
|---|---|
| Output | dictionary. noun.0 |

Table 2: Input and output sequence.

For the approach with KB, the training data has been utilized and arranged in a trie structure based on the POS tag and the word. The word is taken as a root node while POS tags are assigned to first level parent node. All the related instances from the dataset have been stored in the leaf nodes accordingly. The computed tree structure is stored in a JSON file. This structure helps to extract the relevant examples from the KB in a constant time, despite the size of the training set. The training data distribution of the FEWS training set is presented in Table 3.

| POS Tag | No. of Records | POS Tag | No. of Records |
|---|---|---|---|
| Nouns | 55442 | Adjectives | 19269 |
| Verbs | 24396 | Adverbs | 2324 |
| Total | | | 101458 |

Table 3: FEWS dataset distribution.

## 3.2 Optimal prompt selection using prompt augmentation

The study was conducted in three main phases inspired from our previous work (D. Sumanathilaka et al., 2024). The first phase aimed to identify the optimal prompt for extracting the correct sense ID from the sense tags associated with ambiguous words within a given sentence. This phase employed a human-in-the-loop approach, where the lead researcher used prompt engineering techniques to develop the most suitable prompt for extracting the sense ID. An iterative approach was adopted, with careful refinement of the prompt based on the results of each iteration. Incorrect predictions were systematically analyzed to improve the prompt and generate optimal results. This phase explored various prompting techniques, including zero-shot prompting, few-shot prompting, and COT prompting, to identify the most effective approach. Three notable approaches were benchmarked in Table 5, comparing their results based on POS tag.

The initial phase utilized the GPT-3.5 Turbo model. During this phase, both general zero-shot prompting and zero-shot COT prompting techniques were evaluated. Filtered gloss definitions of the ambiguous words were provided to LLM to identify the correct word sense. However, the results revealed that some challenging ambiguous words could not be identified without a proper understanding of each sense tag. To address this limitation, a KB approach using few-shot COT prompting was proposed to enhance in-context learning. The model was prompted with example cases of each sense tag along with their corresponding glosses. The KB used was created from the training data of the FEWS dataset. The optimal prompt selected for this phase is presented in Table 4.

Within the prompt definition, {filtered_definitions} holds the refined definitions extracted from the FEWS sense tag and includes their corresponding sense IDs. The {sentence} section features the original sentence where ambiguous words are highlighted using <WSD> tags. To facilitate a deeper understanding of each ambiguous word, the {examples} section provides relevant instances from the dataset. A detailed flow is shown in Figure 1.

| Improved prompting with knowledge base |
|---|
| You are going to identify the corresponding sense tag of an ambiguity word in English sentences. Do the following tasks.<br><br>1. {word} has different meanings. Below are possible meanings. Comprehend the sensetags and meanings. {filtered_definitions}<br>2. You can learn more on the usage of each word and the meaning through below Examples. Examples are "{examples}".<br>3. Now examine the sentence below. You are going to identify the most suitable meaning for ambiguity word. "{sentence}"<br>4. Try to identify the meaning of the word in the above sentence which is enclosed with the <WSD>. You can think of the real meaning of sentence and decide the most suitable meaning for the word.<br>5. Based on the identified meaning, try to find the most appropriate senseIDs from the below. You are given definition of each sense tag too."{filtered_definitions}".<br>6. If you have more than one senseIDs identified after above steps, you can return the senseIDs in order of confidence level.<br>7. Return JSON object that contains the ambiguity word and the finalized senseIDs.<br>Use the following format for the output. <JSON Object with ambiguity word and the finalized senseIDs > |

Table 4: Optimal selected prompt after phase 1.

## 3.3 Commercial and open-Source model evaluation phase

There are few evaluation techniques proposed in the literature to identify the LLM capabilities on contextual understanding (Guo et al., 2023). Among them, GLUE (A. Wang et al., 2018) and SuperGLUE (A. Wang et al., 2019) are considered to be frequently used evaluation metrics. These techniques are mainly focused on evaluating diverse NLP tasks. However, there prevails a requirement to have a proper matrix for LLM evaluation to benchmark the language understanding when ambiguity exists in a natural text. This phase introduces a proper pipeline for the WSD evaluation for LLMs with the few-shot COT prompting technique. The optimal prompt identified by Phase 1 has been used to conduct the Phase 2 study. The benchmark of the base models is performed using the testing data consisting of 1050 data instances grouped according to the POS tag. The experiment set up for each model evaluation and the results have been discussed in subsection 3.5 and Table 6 respectively. The evaluation was mainly conducted in 2 directions namely evaluation of performance as a prediction model considering the highest confidence answer and the evaluation of performance as a suggestion model. For the suggestion model, the two most confident sense tag predictions were considered.

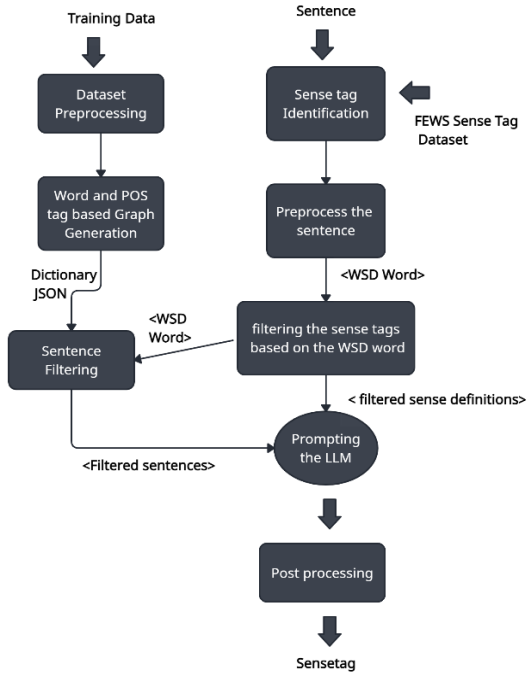

Figure 1: Data flow of the proposed approach.

### 3.4 Parameter and prompt tuning on corner cases

This phase of the study is mainly focused on improving the performance of the module by adding external parameters and different prompt tuning techniques on the incorrectly predicted instances from the study in phase 1. Three approaches were suggested and evaluated using GPT 3.5 and GPT 4 Turbo models as the base models. The selection of the GPT for the third study is motivated due to its performance during phase 1. Self-consistency (X. Wang et al., 2023) prompting was used with the majority voting to decide the final result using a multiple reasoning approach. Steps 4 and 5 of the optimal prompts (See Table 5) were amended. As proposed by previous works, synonyms enhance the learning space on the gloss of an ambiguous word. Therefore, during the next approach synonyms of each WSD word have been shared along with sense ID to enhance the lexical knowledge (Li et al., 2023). This has helped the model to learn more insights about the gloss of the sense IDs. In the last approach, prompt chaining has been incorporated with an aspect-based filtering method. The initial prompt was assigned to filter the sense tags based on the aspect of the sentence. The filtered sense ID has been shared with the second prompt for predicting the final sense ID. Updated prompts for the aspect-based filtering approach can be found in the appendix.

### 3.5 Experimental setup

In this study, our objective was to assess the effectiveness of various prompting strategies using widely utilized LLMs. We have chosen flagship models from leading LLM providers for the study based on their accessibility. GPT 3.5 Turbo, GPT-4-0125-preview models (available by April 2024), Virtuoso-Large by Arcee.ai and Gemini models were chosen due to proven expertise in various languages understanding tasks (Guo et al., 2023). We obtained an OpenAI API key from a tier-one OPEN AI account to access the model and Gemini API key from Google API Studio. During the evaluation process, the model was configured to maintain a temperature of 0 and a maximum token limit of 500 for each output. The Open source LLMs were accessed through 'together' API maintaining the same temperature and the token limit. Deepseek was inferenced through openrouter. The primary task assigned to all the LLMs was word sense identification defining their role as "helpful assistant for identifying word senses". To conduct our evaluation, we utilized test data sourced from the FEWS dataset. The selection of this data followed a 4:3:3 ratio for nouns, verbs, and adjectives, respectively. Additionally, we evaluated 50 instances of adverbs. Overall, our evaluation set comprised 1050 instances. For each testing instance, we considered disambiguation to be correct if the predicted sense tag aligned with the target sense tag. Accuracy was calculated as the percentage of correct predictions relative to all test cases. If more than one sense tag is identified through the models, it has been ordered to the confident values and analysis has been done accordingly (refer to Table 6). Subsequently, we analyzed the number of correct predictions, execution time, and token distribution to assess the model's performance.

## 4 Results and discussion

Table 5 shows the accuracy and statistically significant differences of each approach evaluated in phase 1 to identify the optimal prompting. The study shows that the KB approach with enhanced prompting outperforms the WSD predictions compared to the human-centered general prompts. The few shots provided during the process have enhanced lexical usability and the pragmatic relationship of each ambiguous word. The approach has enriched the lexical knowledge for the inference process. Notably, nouns outperform other POS tags in all the approaches because nouns often behave as concrete concepts with less ambiguity. The optimal prompt used a high average execution time and token amount compared to other general approaches. The results highlighted in bold show the best results while * indicates the statistical significance of the best performance model compared to all the other approaches. The McNemar tests were conducted against each approach and the statistical significance ($P < 0.05$) is noted in Table 5. All word accuracy consists of nouns, verbs, adjectives and adverbs.

| Approach | POS Tag | | |
|---|---|---|---|
| | Noun | Verb | All Word |
| | 400 | 300 | 1050 |
| General prompting with a knowledge base | 0.70 | 0.60 | 0.65 |
| Enhanced prompting with knowledge base | 0.76 | 0.65 | 0.70 |
| Improved prompt with prompt augmentation | **0.85*** | **0.78*** | **0.82*** |
| * Indicate statistically significant differences (p < 0.05) using a McNemar test | | | |

Table 5: Results of optimal prompts.

Table 6 presents the results for the performance of each LLM for the WSD task. The accuracy of each module is presented with suggestion level (S) and prediction level (P), respectively. The improved prompt with prompt augmentation from phase 1 is used to evaluate all the LLMS during this phase (refer to Table 4). Models used for the study are chat or instruct-tuned models of each LLM.

| Model | | POS Tag | | | | |
|---|---|---|---|---|---|---|
| | | Noun | Verb | Adj | Adv | All word |
| | | 400 | 300 | 300 | 50 | 1050 |
| Deepseek R1 | P | **0.91** | **0.83** | **0.87** | **0.88** | **0.88** |
| O4-mini | P | 0.89 | 0.82 | 0.84 | 0.82 | 0.85 |
| Virtuoso-Large | S | **0.91** | **0.84** | **0.90** | **0.86** | **0.89** |
| | P | 0.85 | 0.79 | 0.84 | 0.84 | 0.83 |
| GPT 4 Turbo | S | 0.86 | 0.77 | 0.82 | 0.78 | 0.82 |
| | P | 0.85 | 0.77 | 0.82 | 0.78 | 0.81 |
| GPT 3.5 Turbo | S | 0.85 | 0.78 | 0.80 | 0.86 | 0.82 |
| | P | 0.81 | 0.75 | 0.75 | 0.76 | 0.79 |
| Llama-3 70B | S | 0.85 | 0.75 | 0.81 | 0.72 | 0.80 |
| | P | 0.83 | 0.71 | 0.78 | 0.72 | 0.77 |
| GPT 4.1 mini | P | 0.79 | 0.72 | 0.78 | 0.78 | 0.76 |
| Llama-2 70B | S | 0.88 | 0.79 | 0.84 | 0.84 | 0.83 |
| | P | 0.67 | 0.56 | 0.61 | 0.58 | 0.61 |
| Gemini 2.0 Flash | P | 0.79 | 0.67 | 0.76 | 0.66 | 0.74 |
| Gemini 1.5 Flash | S | 0.77 | 0.63 | 0.73 | 0.74 | 0.74 |
| | P | 0.76 | 0.63 | 0.73 | 0.74 | 0.71 |
| Yi - 34B | S | 0.80 | 0.66 | 0.75 | 0.74 | 0.76 |
| | P | 0.65 | 0.51 | 0.57 | 0.52 | 0.58 |
| Gemma 7B | S | 0.73 | 0.65 | 0.73 | 0.76 | 0.73 |
| | P | 0.49 | 0.41 | 0.51 | 0.46 | 0.47 |
| Mixtral 7B | S | 0.68 | 0.61 | 0.73 | 0.80 | 0.70 |
| | P | 0.43 | 0.32 | 0.46 | 0.42 | 0.41 |
| GPT 4o-mini | S | 0.37 | 0.30 | 0.31 | 0.32 | 0.33 |

S: Suggestion level (Most 2 confident answers), P: Prediction level (Best answer)

Table 6: LLM evaluation for WSD.

The results of phase 2 present a comprehensive analysis of disambiguation techniques applied to the challenging task of WSD. Two distinct approaches were explored: suggestion level assessment, focusing on the most confident predictions among multiple sense tags, and prediction level assessment, prioritizing the single most confident sense tag. The suggestion level approach of WSD is important in response generation and information retrieval applications, while prediction level models can be integrated with translation and transliteration systems. Notably, Virtuosa Large by Arcee.ai

exhibits promising performance in suggestion level disambiguation, whereas Deepseek R1 and GPT 4 Turbo outperform in prediction level accuracy. Llama-3-70B, which is an open-source model, shows promising results in prediction level though it is not capable of surpassing the results of GPT-4-Turbo model. Furthermore, comparative analysis of POS tag distributions across all studies reveals nouns and adjectives as relatively easier to disambiguate, whereas verbs need further investigation for enhanced accuracy. These findings offer valuable insights into optimizing WSD techniques across diverse linguistic contexts. Table 7 represents the result of phase 3. The instances not correctly identified by GPT 3.5 turbo and GPT 4 in the prediction level were extracted for the next phase of the study. These false predictions were evaluated with different prompt enhancements and parameter tuning methods. A significant improvement in the results is seen in the proposed approaches. The study was conducted on false predictions of GPT models with improved prompt. 234 instances were evaluated with GPT 3.5 and 191 instances with GPT 4.

| Approach | GPT 3.5 | | GPT 4 | |
|---|---|---|---|---|
| Prompting with self-consistency prompting with a majority vote | 57 | 0.24 | **54** | **0.28** |
| Incorporating synonyms with the prompt | **68** | **0.29** | 42 | 0.21 |
| Incorporating prompt chaining with aspect-based sense filtering and synonyms | 49 | 0.20 | **58** | **0.30** |

Table 7: Results of parameter and prompt tuning.

Phase 3 of the study showcases the efficacy of prompt chaining coupled with aspect-based sense filtering enhanced by synonyms, yielding remarkable results in experimentation. The incorporation of synonyms notably enriches the contextual understanding within the reasoning process, showcasing the potential of this approach to augment WSD tasks. The self-consistency approach which uses multiple reasoning strategies with majority vote shows promising results with GPT 4 while sense space reduction approach with aspect-based filtering shows a new avenue to improve the WSD. Utilizing different computational techniques, we successfully disambiguated some edge cases that had previously posed challenges during the initial studies.

However the observed improvements may appear relatively modest, they highlight a promising direction for future research aimed at refining WSD methodologies. This innovative methodology not only highlights the importance of context in disambiguation but also suggests avenues for further enhancement in the pursuit of more accurate and nuanced disambiguation techniques.

## 5 Conclusion and future directions

This research illustrates the effectiveness of integrating prompt augmentation techniques using large language models with a knowledge-driven strategy to address word sense ambiguity. Future work should focus on evaluating these techniques on comprehensive datasets such as Semcor, SenseEval, and SemEval to provide a robust validation of their efficacy. Additionally, exploring the potential for enhancing performance through the incorporation of additional parameters warrants further investigation. By accurately disambiguating the true meaning of words within their context, we can significantly enhance Cyber Threat Intelligence efforts, thereby curbing the spread of misinformation in natural text.

This study introduces a novel method for Word Sense Disambiguation that incorporates prompt augmentation within a human-in-the-loop framework, yielding promising results and suggesting practical utility for various NLP-based tasks. Subsequent research will aim to expand this approach across a diverse array of fine-tuned commercial and open-source models, to validate its generalizability and explore its applicability across various real-world scenarios. This comprehensive approach not only advances the state-of-the-art in WSD but also opens new avenues for practical applications in natural language processing.

**Acknowledgements**

We extend our heartfelt gratitude to the reviewers for their insightful comments and constructive feedback, which significantly improved the quality of this paper. We also thank the OpenAI Researcher Access Program for providing credits to support this project's development.

# Appendix

The Table 8 contains the prompts used for phase 3 of the study.

| Enhanced prompting |
|---|
| You are going to identify the sense tag of an ambiguity word in English. <br><br> Do the following tasks. <br> 1. Examine the sentence below. "{sentence}". <br> 2. Identify the meaning of the word enclosed within the <WSD> tags. You need to consider the total sentence before you get the exact meaning of the word. <br> 3. Based on the identified meaning, try to find the most appropriate senseIDs from the below. "{meanings}". <br> 4. If you have more than one senseIDs identified, you can return the senseIDs in order of confidence level. <br> 5. Return a proper JSON object that contains the ambiguity word and the finalized senseIDs. <br> Use the following format for the output. <br> <JSON object that contain ambiguity word and the finalized senseIDs> |
| **Self-consistency prompt [ 1st approach study 3.4]** |
| You are going to identify the corresponding sense tag of an ambiguous word in English sentences. Use multiple reasoning strategies to increase confidence in your answer. <br> 1. The word "{wordwsd}" has different meanings. Below are possible meanings. Comprehend the sense tags and meanings: {filtered_definitions} <br> 2. You can learn more on the usage of each word and the meaning through the examples below. Each sentence is followed by its corresponding sense id. "{examples}" <br> 3. Now carefully examine the sentence below. The ambiguous word is enclosed within <WSD>."{sentence}" <br> 4. Analyze the sentence using the following three approaches. For each approach, identify the meaning of the ambiguous word and the corresponding sense IDs. If there are multiple sense IDs, separate them with commas. <br> Strategy 1: Focus on keywords in the sentence surrounding the ambiguous word. See which sense definition aligns best with these keywords. <br> Strategy 2: Consider the part of speech (noun, verb, adjective, etc.) of the ambiguous word in the sentence and how it functions within the sentence structure. Choose the sense definition that fits this grammatical role. <br> Strategy 3: Think about the overall topic and intent of the sentence. Decide on the sense of the word that makes the most logical sense within the wider context. <br> 5.Compare the sense ID(s) identified by each strategy. <br> If all three strategies agree on the same sense ID, that is your most confident answer. <br> If two strategies agree on a same sense ID, that becomes your answer. <br> If there is a disagreement, list the sense ID(s) from each strategy for further review. <br> 6. Return a JSON object containing the following: <br> "word": The ambiguous word <br> "sense_id": The sense ID(s) determined as most likely based on the majority vote <br> "strategy_1": Sense ID(s) suggested by Strategy 1 <br> "strategy_2": Sense ID(s) suggested by Strategy 2 <br> "strategy_3": Sense ID(s) suggested by Strategy 3 <br> ''' |

| Prompt tuning with synonyms [2nd approach study 3.4] |
|---|
| You are going to identify the corresponding sense tag of an ambiguous word in English sentences. Use multiple reasoning strategies to increase confidence in your answer.<br>1. The word "{wordwsd}" has different meanings. Below are possible meanings. Comprehend the sense tags and meanings. Synonyms are provided if available. {filtered_definitions}<br>2. You can learn more on the usage of each word and the its sense through the examples below. Each sentence is followed by its corresponding sense id. "{examples}"<br>3. Now carefully examine the sentence below. The ambiguous word is enclosed within <WSD>."{sentence}"<br>4. Analyze the sentence using the following techniques and identify the meaning of the ambiguous word.<br>  Focus on keywords in the sentence surrounding the ambiguous word.<br>  Think about the overall topic and intent of the sentence. Decide on the sense of the word that makes the most logical sense within the context.<br>5. Based on the identified meaning, try to find the most appropriate senseIDs from the below sense tag list. You are given definition of each sense tag too."{filtered_definitions}".<br>6. If you have more than one senseIDs identified after above steps, you can return the senseIDs in order of confident level, follow the given format to return the value.<br>7. Return a JSON object containing the following: "word": The ambiguous word, "sense_id": The sense ID(s) ' |

| Prompt chaining with aspect-based sense filtering [3rd approach of the study 3.4] |
|---|
| *Prompt 1:*<br>You are going to identify the corresponding sense tags of an ambiguous word in English sentences. Use multiple reasoning strategies to increase confidence in your answer.<br>1. The word "{wordwsd}" has different meanings. Below are possible meanings. Comprehend the sense tags and meanings. Synonyms are provided if available. {filtered_definitions}<br>2. Now carefully examine the sentence below. The ambiguous word is enclosed within <WSD>."{sentence}"<br>4. Analyze the sentence using the following techniques and identify the appropriate sense tags of the ambiguous word.<br>   -Focus the aspect discussed in the above sentence and filter the relevant sense tags.<br>   -Think about the overall topic and intent of the sentence. Decide on the sense tags of the word that makes the most logical sense within the context.<br>5. Now you can return all sense IDS identified by the above steps.<br>7. Return a JSON object containing the following:<br>   <"sense_id": The sense ID(s), "sense meaning": Summarized Sense meaning ><br><br>*Prompt 2:*<br>You are going to identify the corresponding sense tag of an ambiguous word in English sentences.<br>1. The word "{wordwsd}" has different meanings. Below are possible meanings. Comprehend the sense tags and meanings. {definitions}<br>2. You can learn more on the usage of each word and its sense through the examples below if provided. Only focus on the sentences with above sensetags. You can discard sentences with different sense tags.Each sentence is followed by its corresponding sense id. "{examples}"<br>3. Now carefully examine the sentence below. The ambiguous word is enclosed within <WSD>."{sentence}"<br>4. Analyze the sentence using the "keywords surrounding the ambiguous word" and the "overall topic and meaning of the sentence" and identify the meaning of the ambiguous word.<br>5. Based on the identified meaning, try to find the most appropriate senseID (only one) from the below sense tag list. You are given definition of each sense tag too."{definitions}".<br>6. Return a JSON object containing the following: "word": The ambiguous word, "sense_id": The sense ID |

Table 8: Prompts used for handling corner cases.